\documentclass{article}
\usepackage{float}
\usepackage{amsmath}
\usepackage{PRIMEarxiv}
\usepackage{makecell}
\usepackage[utf8]{inputenc} 
\usepackage[T1]{fontenc}    
\usepackage{hyperref}       
\usepackage{url}            
\usepackage{booktabs}       
\usepackage{amsfonts}       
\usepackage{nicefrac}       
\usepackage{microtype}      
\usepackage{lipsum}
\usepackage{fancyhdr}       
\usepackage{graphicx}       
\graphicspath{{media/}} 
\usepackage{xcolor}
\usepackage{authblk}
\usepackage{hyperref}

\begin{document}

\title{A Novel Ophthalmic Benchmark for Evaluating Multimodal Large Language Models with Fundus Photographs and OCT Images}
\author[1,3,†]{Xiaoyi Liang}          
\author[2†]{Mouxiao Bian}      
\author[1,3,†]{Moxin Chen}     
\author[2]{Lihao Liu}
\author[2]{Junjun He}
\author[2,*]{Jie Xu}
\author[1,3,*]{Lin Li}
 
\affil[1]{\textit{
    Ninth People’s Hospital, Shanghai Jiao Tong University School of Medicine \\
    Shanghai, China
}}
 
\affil[2]{\textit{
    Shanghai Artificial Intelligence Laboratory, \\
    Shanghai, China
}}
 
\affil[3]{\textit{
     Shanghai Key Laboratory of Orbital Diseases and Ocular Oncology\\
    Shanghai, China
}}
 
\footnotetext[1]{†These authors contributed equally.}
\footnotetext[2]{*Correspondence: 
Lin Li (lin\_li@sjtu.edu.cn); 
Jie Xu (xujie@pjlab.org.cn)
}

\maketitle

\begin{abstract}
In recent years, large language models (LLMs) have demonstrated remarkable potential across various medical applications. Building on this foundation, multimodal large language models (MLLMs) integrate LLMs with visual models to process diverse inputs, including clinical data and medical images. In ophthalmology, LLMs have been explored for analyzing optical coherence tomography (OCT) reports, assisting in disease classification, and even predicting treatment outcomes. However, existing MLLM benchmarks often fail to capture the complexities of real-world clinical practice, particularly in the analysis of OCT images. Many suffer from limitations such as small sample sizes, a lack of diverse OCT datasets, and insufficient expert validation. These shortcomings hinder the accurate assessment of MLLMs' ability to interpret OCT scans and their broader applicability in ophthalmology.
To address this issue, we introduce a novel multimodal benchmark for ophthalmology that leverages fundus photographs and OCT images to comprehensively evaluate the diagnostic performance of MLLMs. Our dataset, curated through rigorous quality control and expert annotation, consists of 439 fundus images and 75 OCT images. Using a standardized API-based framework, we assessed seven mainstream MLLMs and observed significant variability in diagnostic accuracy across different diseases. While some models performed well in diagnosing conditions such as diabetic retinopathy and age-related macular degeneration, they struggled with others, including choroidal neovascularization and myopia, highlighting inconsistencies in performance and the need for further refinement. Our findings emphasize the importance of developing clinically relevant benchmarks to provide a more accurate assessment of MLLMs' capabilities. By refining these models and expanding their scope, we can enhance their potential to transform ophthalmic diagnosis and treatment.

\end{abstract}


\section{Introduction}

Large language models (LLMs) are artificial intelligence models designed to understand and generate human language \cite{yu2023leveraging}. Trained on vast amounts of text data, they can perform a wide range of tasks, including medical education, triage, diagnosis, treatment, and prognosis \cite{zhao2021calibrate, cascella2023evaluating, li2024integrated, jin2024matching, shen2023chatgpt}. Building on this foundation, multimodal large language models (MLLMs) integrate LLMs with large visual models (LVMs) to process a broader range of inputs, such as clinical data, examination results, and medical images \cite{zhou2024pre, bhattacharya2024large, liu2023benchmarking}. In ophthalmology, MLLMs have been explored for analyzing optical coherence tomography (OCT) images, aiding in disease classification, and predicting treatment outcomes. These advancements highlight MLLMs’ potential for real-world clinical applications and their significant research value. To evaluate MLLM performance, several benchmarks have been developed \cite{besler2024accuracy, alryalat2024evaluating, jin2023medcpt, liu2024medbench}. However, existing benchmarks often exhibit key limitations, particularly in assessing OCT interpretation.

First of all, most existing benchmarks have limited clinical utility, as they primarily source questions from major medical exam banks such as the United States Medical Licensing Examination (USMLE) \cite{lee2006beyond, liu2023medical, pal2022medmcqa}. These question banks pose two key issues: (i) their high risk of exposure makes it difficult to assess a model’s true capabilities; (ii) they focus on theoretical knowledge and standardized clinical cases while overlooking the complexities of real-world practice. \textit{Therefore, current benchmarks fail to evaluate models' ability to handle differential diagnosis, atypical cases, and nuanced clinical reasoning.}

Moreover, these benchmarks often lack sufficient discriminatory power \cite{lin2023comparison}. Score differences between models are frequently minimal, making it difficult to gauge their actual ability to address complex clinical problems \cite{healey2025llm, chen2024evaluating}. This challenge is even greater for multimodal models, as the diversity of medical images and variations across imaging devices complicate the establishment of standardized evaluation metrics \cite{ma2022benchmarking, kus2024medsegbench, gong2023pimedseg}. Additionally, privacy concerns and ethical constraints limit access to real-world medical images, making it difficult to curate high-quality, representative datasets for benchmarking \cite{kus2024medsegbench, kocak2025bias, liu2020evolving}. \textit{As a result, existing benchmarks struggle to balance the demands of multimodal evaluation, large-scale data, and rigorous expert annotation.}

To address these challenges, we developed a multimodal benchmark focused on ophthalmology, specifically targeting fundus photographs and optical coherence tomography (OCT) images. Ophthalmology is a highly specialized field where diagnosis heavily relies on imaging \cite{ting2019artificial, costin2025artificial, fujinami2019prediction, li2023generic}. Early disease identification and differential diagnosis require extensive expertise, but access to skilled ophthalmologists is often limited due to disparities in medical resources \cite{baxter2022data, lee2016evaluating}. Given the image-centric nature of ophthalmology, this field provides an ideal setting to explore the potential of multimodal models in clinical diagnosis and decision-making. Using our curated dataset, we evaluated the diagnostic capabilities of several prominent MLLMs in interpreting ophthalmic images. Our benchmark has been integrated into MedBench’s ophthalmology track—Comprehensive Capabilities and Multimodal Assessment for Eye Health Specialties—which provides a standardized framework for assessing model performance in ophthalmic disease diagnosis. More details can be found at https://medbench.opencompass.org.cn/track.

\section{\textbf{\textbf{Methods}}}
\subsection{Evaluation Models}
We selected seven well-known multimodal large language models with image understanding capabilities that can be accessed via API calls, including two models with over a hundred billion parameters and five models with fewer than a hundred billion parameters. Since the QVQ-72B-Preview has hardly correctly identified OCT images, we have not included its results in the analysis. Details can be found in Table 1.
\begin{table}
  \caption{Introduction of MLLMs}
  \centering
  \begin{tabular}{llllll}
    \toprule
    Model & Parameters & Context length & Open source model & Task(s) & \makecell{Cost of Input\\ (\$/1M Tokens)} \\
    \midrule
    InternVL2-8B & 8B & 8K & Yes & FPD, OCTD & / \\
    Gemini-2.0-Flash & 540B & 1M & No & FPD, OCTD & 0.1 \\
    GPT-4o & About 200B & 4K & No & FPD, OCTD & 2.5 \\
    GPT-4o-mini & About 8B & 65K & No & FPD, OCTD & 0.15 \\
    QVQ-72B-Preview & 72B & 32K & Yes & FPD & / \\
    Qwen2.5-VL-72B-Instruct & 72B & 128K & Yes & FPD, OCTD & / \\
    Qwen2-VL-72B-Instruct & 72B & 32K & Yes & FPD, OCTD & / \\
    \bottomrule
  \end{tabular}
  \raggedright
    \textit{Note:FPD, fundus photograph diagnosis; OCTD, optical coherence tomography diagnosis}
\end{table}
\subsection{Consist of Ophthalmic Benchmark}
Our benchmark includes fundus photographs and OCT images(FIgure 1).Fundus photographs are in the following 17 conditions to assess 7 MLLMs: diabetic retinopathy (DR) images, normal (NORMAL) images, media haze (MH) images, optic disc cupping (ODC) images, tessellation (TSLN) images, age-related macular degeneration (ARMD) images, Drusen (DN) images, myopia (MYA) images, branch retinal vein occlusion (BRVO) images, optic disc pallor (ODP) images, central retinal vein occlusion (CRVO) images, choroidal neovascularization (CNV) images, retinitis (RS) images, optic disc edema (ODE) images, laser scars (LS) images, central serous retinopathy (CSR) images and hypertensive retinopathy (HTR) images. And for OCT images, we assessed the diagnostic capabilities of six MLLMs in eight conditions :Age-related macular degeneration(ARMD), choroidal neovascularization(CNV), central serous retinopathy(CSR), diabetic macular edema(DME), diabetic retinopathy(DR), drusen(DRUSEN), macular hole(MH), normal(NORMAL).

\begin{figure}
    \centering
    \includegraphics[width=1\linewidth]{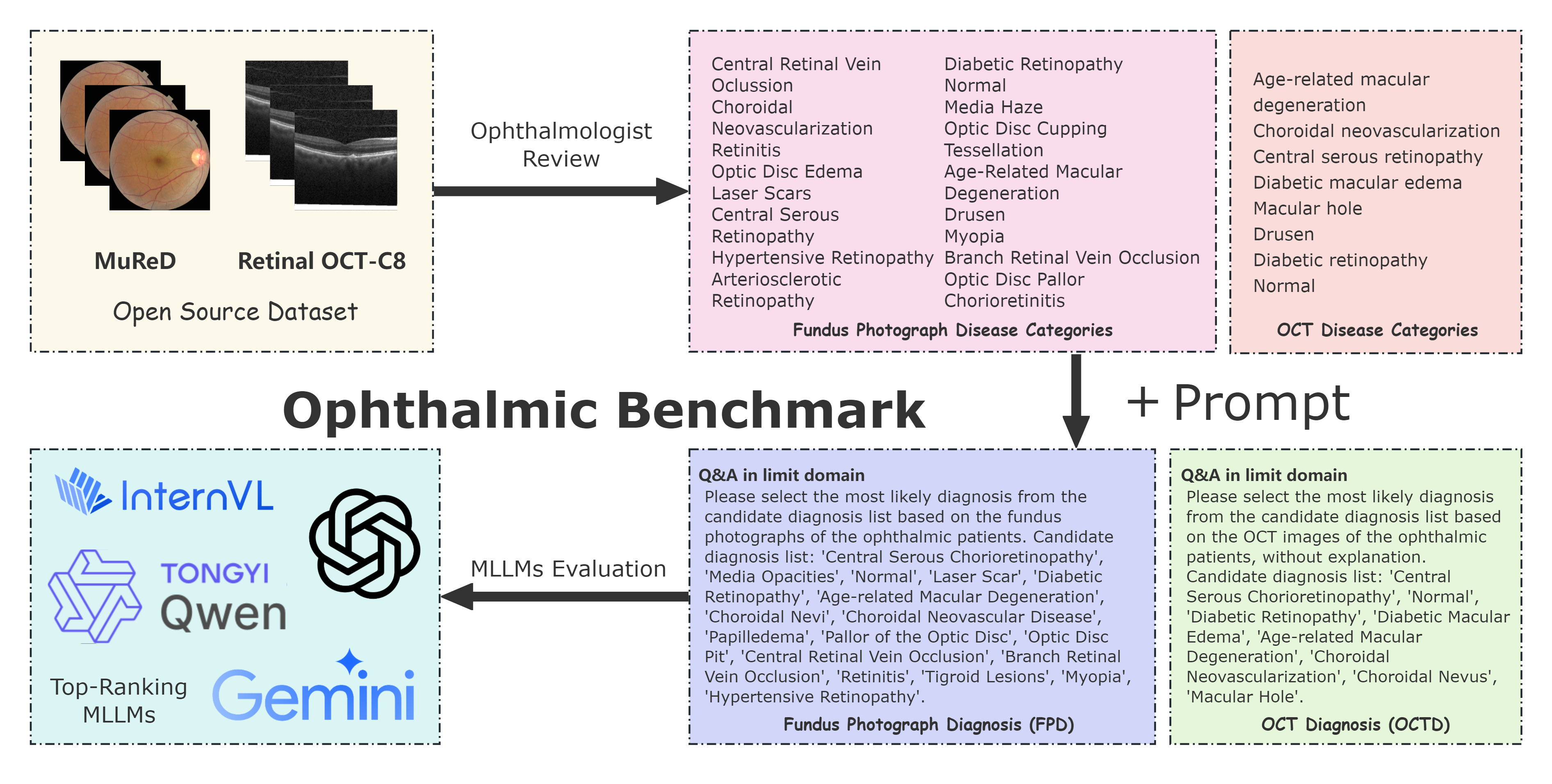}
    \caption{Ophthalmic Benchmark}
    \label{fig:enter-label}
\end{figure}

\section{\textbf{\textbf{Dataset and Experiment}}}

\subsection{Dataset Collection}
The fundus photographic images were obtained from the Multimodal Retinal Dataset (MuReD), a publicly accessible ophthalmic imaging repository, while the cross-sectional retinal layer analyses were derived from optical coherence tomography (OCT) images sourced from Dataset C8 of the Retinal OCT Image Classification database. Both datasets were acquired using clinical-grade ophthalmic imaging systems adhering to standardized imaging protocols, ensuring diagnostic-quality resolution and anatomical structure visualization suitable for computational analysis in retinal pathology research.
\subsection{Data Curation and Multidimensional Annotation}
The original datasets were evaluated by a panel of five ophthalmologists, each with at least 10 years of clinical experience, and subjected to a rigorous triphasic quality control process: (1) Technical validation, which involved the exclusion of illegible images that could not be used for direct diagnosis; (2) Exclusion of images exhibiting advanced comorbidities; (3) Discrepant cases underwent independent verification and consensus-based adjudication by a committee of three senior ophthalmologists (over 15 years' experience), with final label approval contingent upon strict alignment with gold-standard diagnostic criteria. Finally, we prepared a multimodal dataset that has undergone multiple rounds of clinical expert review, consisting of 75 OCT images and 439 fundus images. 
\subsection{Dataset Design for Multimodal Tasks}
Multimodal tasks encompass fundus photograph diagnosis (FPD) and OCT diagnosis (OCTD) to comprehensively evaluate MLLM's capability to diagnose retinal diseases using various imaging modalities. For FPD, the dataset was designed to classify fundus images into different disease categories, such as diabetic retinopathy, retinitis, age-related macular degeneration, and others. For OCTD, the dataset focused on segmenting and analyzing the various retinal layers from OCT images to detect structural abnormalities indicative of disease progression. Each task required the integration of both the fundus photographic and OCT datasets, ensuring a comprehensive and multifaceted evaluation of MLLM's diagnostic accuracy and generalizability across different retinal conditions. Each FPD task is designed as follows:

\textit{\textit{Please select the most likely diagnosis from the candidate diagnosis list based on the fundus photographs of the ophthalmic patients. Candidate diagnosis list: \{'Central Serous Chorioretinopathy', 'Media Opacities', 'Normal', 'Laser Scar', 'Diabetic Retinopathy', 'Age-related Macular Degeneration', 'Choroidal Nevi', 'Choroidal Neovascular Disease', 'Papilledema', 'Pallor of the Optic Disc', 'Optic Disc Pit', 'Central Retinal Vein Occlusion', 'Branch Retinal Vein Occlusion', 'Retinitis', 'Tigroid Lesions', 'Myopia', 'Hypertensive Retinopathy'\}}}

While the OCTD task is as follows:

\textit{\textit{Please select the most likely diagnosis from the candidate diagnosis list based on the OCT images of the ophthalmic patients, without explanation. Candidate diagnosis list: \{'Central Serous Chorioretinopathy', 'Normal', 'Diabetic Retinopathy', 'Diabetic Macular Edema', 'Age-related Macular Degeneration', 'Choroidal Neovascularization', 'Choroidal Nevus', 'Macular Hole'\}.}}

\subsection{Model Validation Framework \& Experimental Design}
We referenced the dataset preparation and evaluation process of the renowned medical large model evaluation platform Medbench, and we tested seven mainstream multimodal large models using the method of calling APIs, uploading images by converting local images into base64 format. By utilizing APIs, we ensured a standardized interface for interacting with these models, allowing for fair comparison based on their performance in diagnosing ophthalmic conditions. The conversion of local images to base64 format facilitated seamless upload and processing, ensuring that the image quality remained consistent across all models. This step was crucial for the accuracy and reliability of our validation framework.
\subsection{Model Performance Evaluation}
The diagnostic accuracy of multimodal models was quantified using a rigorously validated evaluation protocol. Accuracy metrics were calculated as the proportion of correctly classified instances relative to the total test dataset, formulated as:

$$
\text{Accuracy} = \frac{TP + TN}{TP + TN + FP + FN}
$$
where \textit{\textit{TP }}denotes true positives, \textit{\textit{TN}} true negatives,\textit{\textit{ FP}} false positives, and \textit{\textit{FN}} false negatives. 
\subsection{Data analysis and Plotting}
To ensure data consistency and accuracy, we first preprocess the data using the Pandas package in Python, which includes the detection and filtering of outliers, as well as data normalization. Subsequently, we perform statistical analysis using the NumPy package to evaluate the model's performance. Finally, we use the ggplot2 package in R to create heatmaps, rose plots, and bar charts.

\section{\textbf{\textbf{Results}}}

\subsection{Fundus Photograph Diagnosis (FPD)}
We first evaluated the ability of seven multimodal large language models to recognize fundus images. To intuitively illustrate the differences in model performance, we normalized the accuracy scores. The model with the highest accuracy was assigned a score of 90, and the scores of other models were calculated proportionally, as shown in Figure 2. Overall, Gemini-2.0-Flash performs the best compared to other models, followed by Qwen2.5-VL-72B-Instruct. The accuracy of each model in identifying fundus images for normal condition and each retinal disease is shown in Figure 3. None of the seven multimodal large models, including GPT-4o, answered correctly in the four diseases of HTR, LS, ODP and ODC. The accuracy in CNV and MYA identification and diagnosis is also close to 0. In the diagnosis of DR, the accuracy of three models exceeded 60\%. GPT-4o-mimi had the highest accuracy in ODE diagnosis, reaching 73\%. Gemini-2.0-Flash achieved an accuracy of 70\% in ARMD diagnosis, and Qwen-2.5-VL-72B-Instruct reached 70\% in DR diagnosis. These results indicate that while some models demonstrate limited proficiency in specific disease diagnoses, none of the multimodal large models tested showed consistent high accuracy across all diseases. The low accuracy in identifying and diagnosing conditions such as CNV and MYA suggests that these models may not yet be reliable for clinical use in diagnosing these particular diseases. However, the higher accuracy in diagnosing DR, ODE, and ARMD in some models indicates potential for further refinement and improvement in this area. Continued research and development are needed to enhance the diagnostic capabilities of these models, potentially through the integration of more specialized medical knowledge and training data.

\begin{figure}
    \centering
    \includegraphics[width=0.75\linewidth]{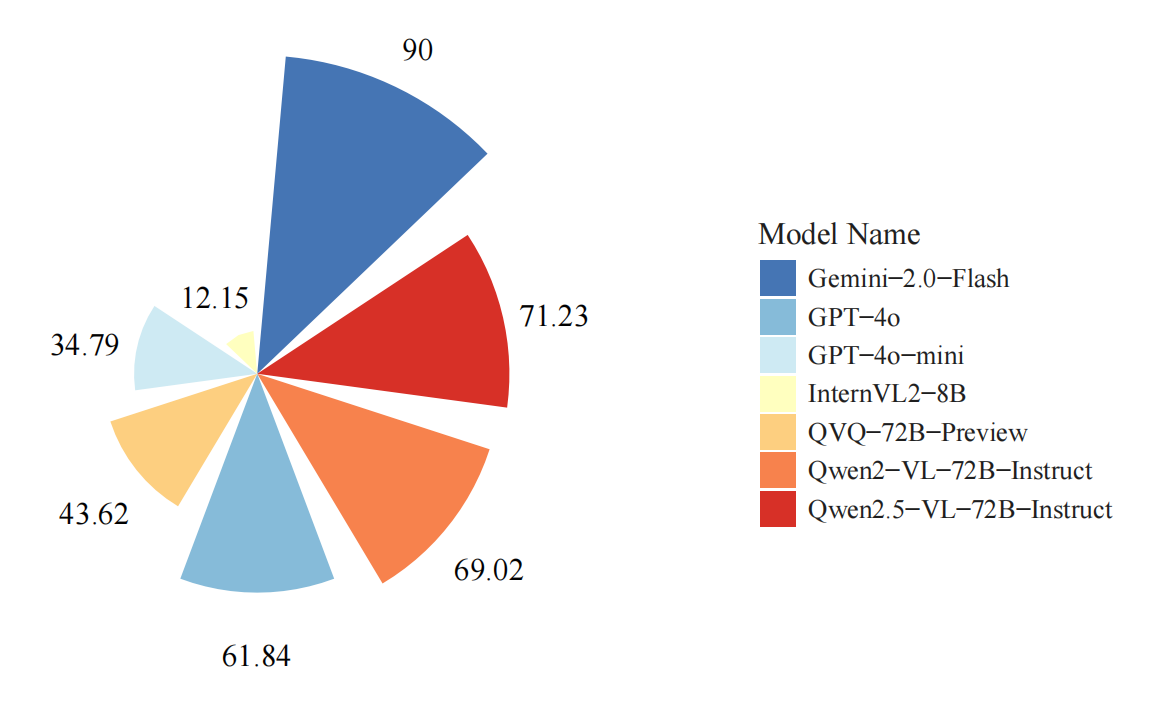}
    \caption{Normalized Accuracy of MLLMs in FPD}
      \raggedright
    \textit{Note:The wreath chart displays the standardized scores of seven models in diagnosing fundus photographs. The model with the highest accuracy received a score of 90, with the scores of the other models determined proportionally. MLLMs, multimodal large language models. FPD, fundus photograph diagnosis.}
    
\end{figure}
\begin{figure}
    \centering
    \includegraphics[width=1\linewidth]{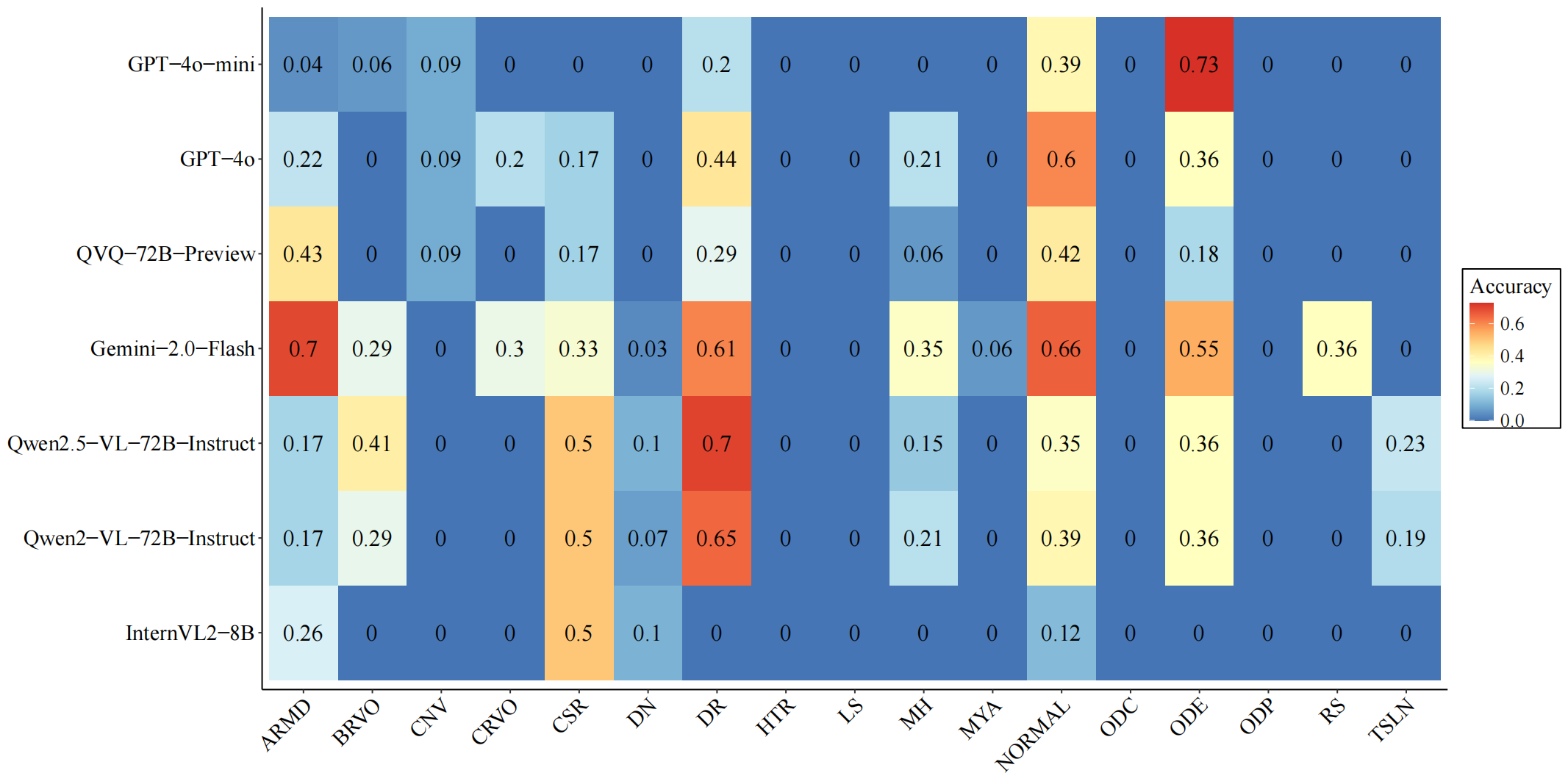}
    \caption{Accuracy of MLLMs in FPD by diseases}
        \raggedright
    \textit{Note:The heatmap illustrates the accuracy of seven models in diagnosing fundus photographs. MLLMs, multimodal large language models. FPD, fundus photograph diagnosis. ARMD, age-related macular degeneration. BRVO, branch retinal vein occlusion. CNV, choroidal neovascularization. CRVO, central retinal vein occlusion. CSR, central serous retinopathy. DN, Drusen. DR, diabetic retinopathy. HTR, hypertensive retinopathy. LS, laser scars. MH, media haze. MYA, myopia. ODC, optic disc cupping. ODE, optic disc edema. ODP, optic disc pallor. RS, retinitis. TSLN, tessellation.}
\end{figure}

\subsection{OCT Diagnosis (OCTD)}
In the OCT diagnosis (OCTD) task, Gemini-2.0-Flash remains the most outstanding model, with GPT-4o and Qwen2-VL-72B-Instruct closely following, both achieving the same score (Figure 4). In Figure 5, similarly, in the identification of CNV disease, the accuracy of all MLLMs' responses remains close to 0. The identification rate for DRUSEN disease is also close to 0. GPT-4o-mini appears to only identify CSR disease, with almost all other disease diagnoses being 0. Gemini-2.0-Flash achieved a 100\% identification accuracy rate for MH, and Qwen2-VL-72B-Instruct also reached 100\% in the identification of MH. This suggests that, while the models tested have limitations in identifying certain diseases, there is potential for improvement in specific areas. The high identification accuracy of MH by Gemini-2.0-Flash and Qwen2-VL-72B-Instruct indicates that with further refinement and training, these models may become more reliable in diagnosing this particular disease. However, the consistent low accuracy in identifying CNV and DRUSEN diseases across all models highlights the need for continued research and development to enhance the diagnostic capabilities of these models. Overall, while some models demonstrate promise in specific areas, none of the multimodal large models tested have shown consistent high accuracy across all diseases in the OCTD task.
\begin{figure}
    \centering
    \includegraphics[width=0.75\linewidth]{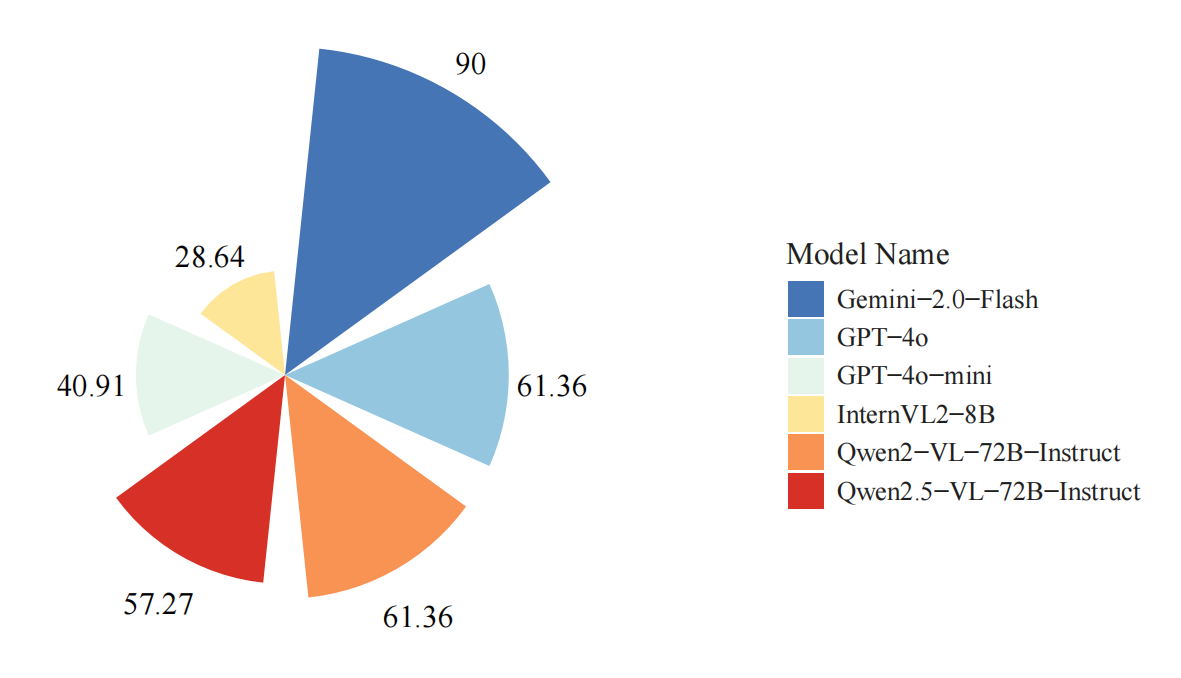}
    \caption{Normalized Accuracy of MLLMs in OCTD}
    \raggedright
    \textit{Note: The wreath chart displays the standardized scores of six models in diagnosing OCT images. The model with the highest accuracy received a score of 90, with the scores of the other models determined proportionally. MLLMs, multimodal large language models. OCTD, optical coherence tomography diagnosis. }
    \end{figure}
\begin{figure}
        \centering
        \includegraphics[width=0.92\linewidth]{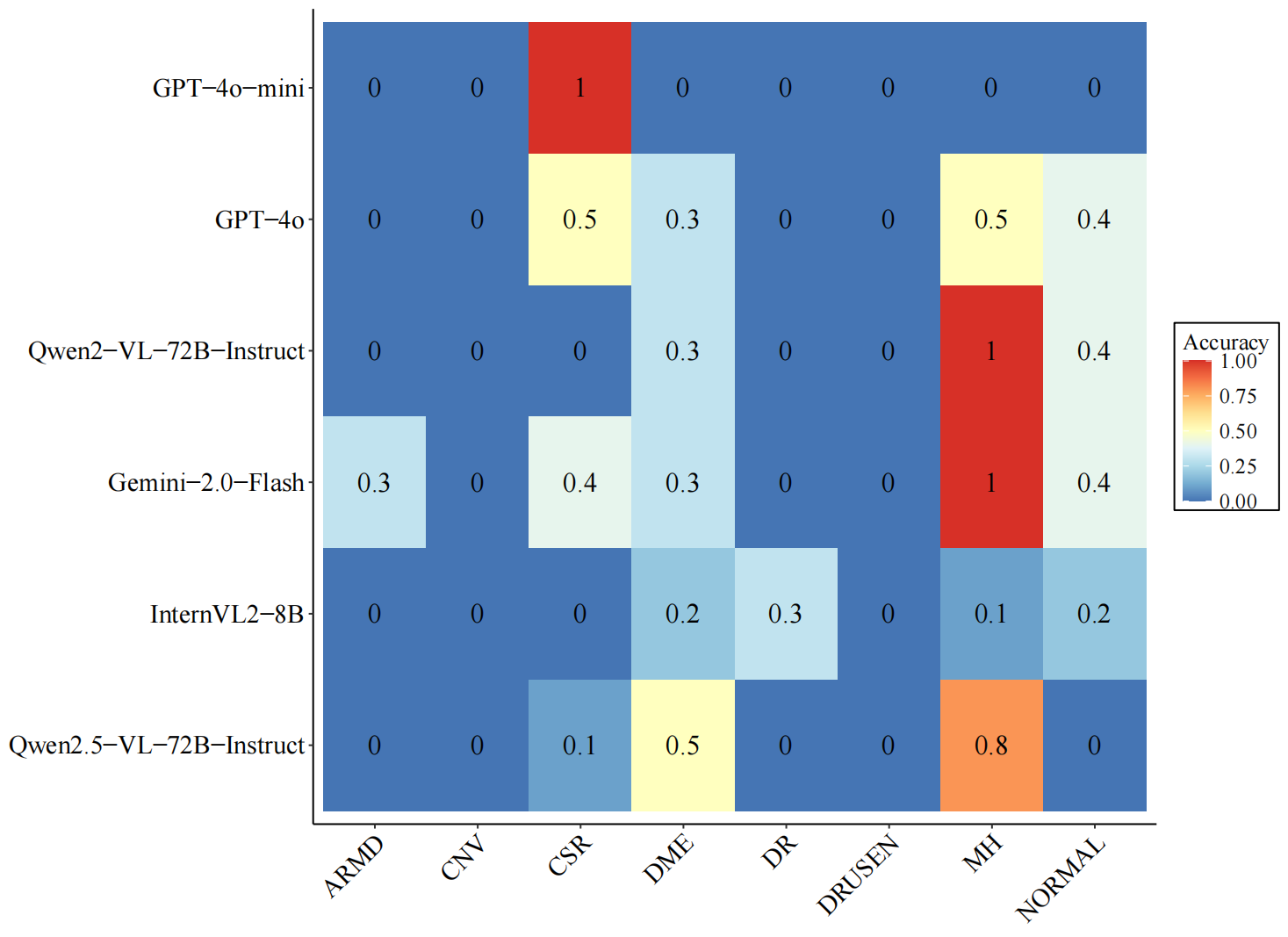}
        \caption{Accuracy of MLLMs in OCTD by diseases}
\raggedright
   Note: \textit{The heatmap illustrates the accuracy of six models in diagnosing OCT images. MLLMs, multimodal large language models. OCTD, optical coherence tomography diagnosis. ARMD, age - related macular degeneration. CNV, choroidal neovascularization. CSR, central serous retinopathy. DME, diabetic macular edema. DR, diabetic retinopathy. MH, media haze.}
    \end{figure}

\subsection{Comparison of Open Source and Non-Open Source Models}
In FPD and OCTD tasks, open-source large language models (InternVL2-8B, QVQ-72B-Preview, and Qwen2-VL-72B-Instruct) have a higher average accuracy than non-open-source large language models (Gemini-2.0-Flash, GPT-4o, GPT-4o-mini, and Qwen2-VL-72B-Instruct), but the gap does not seem to be that significant (Figure 6). This could be due to the fact that open-source models typically have more developers and researchers involved, allowing for faster iteration and improvement. At the same time, open-source models are more accessible and easier to use, which promotes more experimentation and research, thereby enhancing the performance of the models. However, non-open-source models also have their advantages, such as potentially having more proprietary data and algorithms, as well as more specialized development teams. Therefore, when choosing a model, it is necessary to weigh the specific application scenarios and requirements. In the future, as technology continues to evolve, the performance gap between open-source and non-open-source models may further narrow, and more innovative models and methods will emerge, providing more choices and possibilities for the diagnosis and treatment of eye diseases.
\begin{figure}
    \centering
    \includegraphics[width=0.88\linewidth]{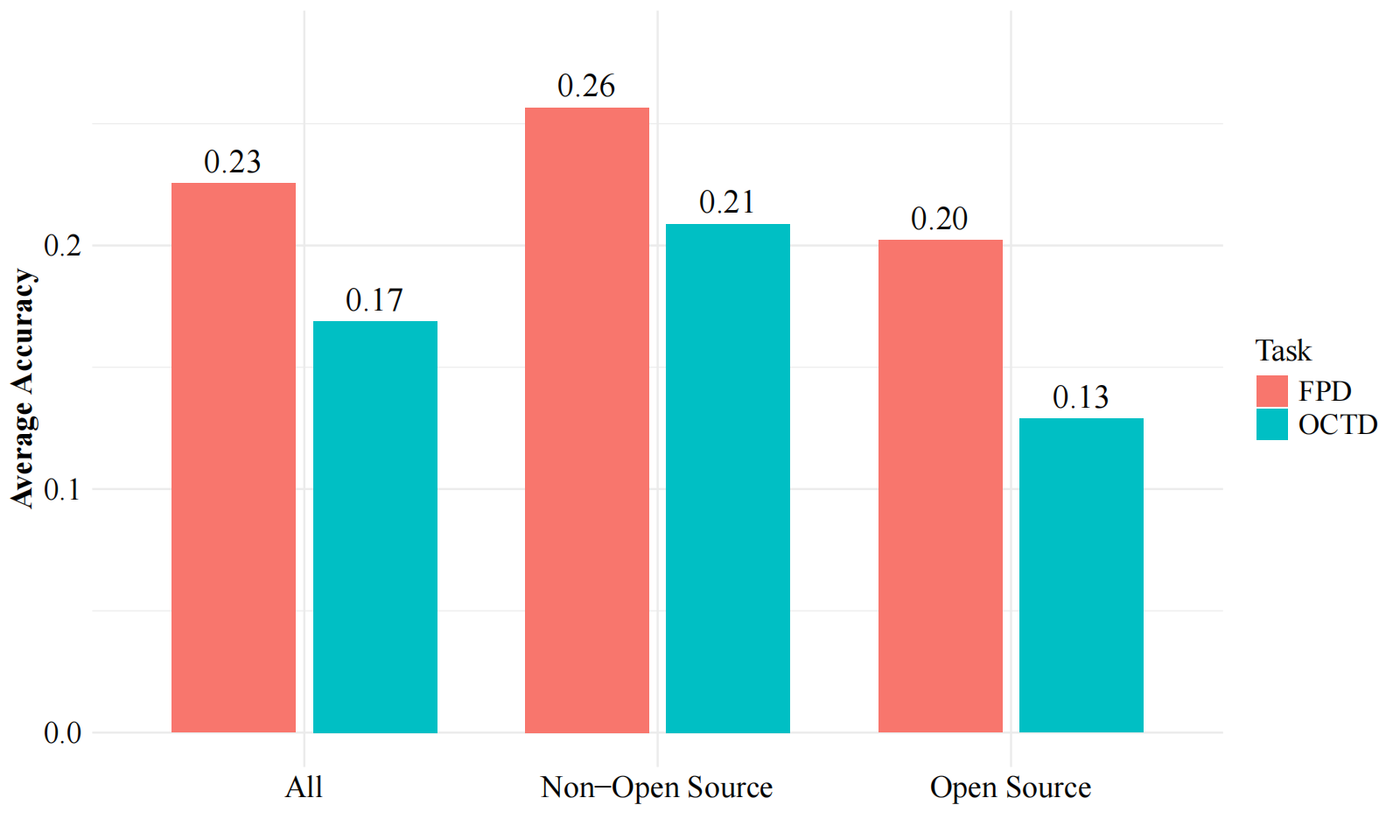}
    \caption{Average Accuracy of Open Source and Non-Open Source MLLMs in FPD and OCTD. }
    \raggedright
   Note: \textit{The bar chart illustrates the performance differences between open-source and non-open-source models on FPD and OCTD tasks. MLLMs, multimodal large language models. FPD, fundus photograph diagnosis. OCTD, optical coherence tomography diagnosis.}
\end{figure}
\subsection{Comparison of Different Parameters Models}
Large language models with over 100 billion parameters tend to have a higher average accuracy on two tasks compared to models with less than 100 billion parameters (Figure 7). However, in the specific case of the FPD task, the accuracy of Qwen2.5-VL-72B-Instruct still surpasses that of GPT-4o (approximately 200 B). This indicates that although the number of parameters is an important factor, it is not the only determinant of model performance. The architecture of the model, the quality of the training data, and optimization for specific tasks can all significantly impact model performance. Therefore, when selecting a model, it is necessary to consider factors beyond the number of parameters, such as the applicability of the model, the availability of training data, and the interpretability of the model.
\begin{figure}
    \centering
    \includegraphics[width=0.88\linewidth]{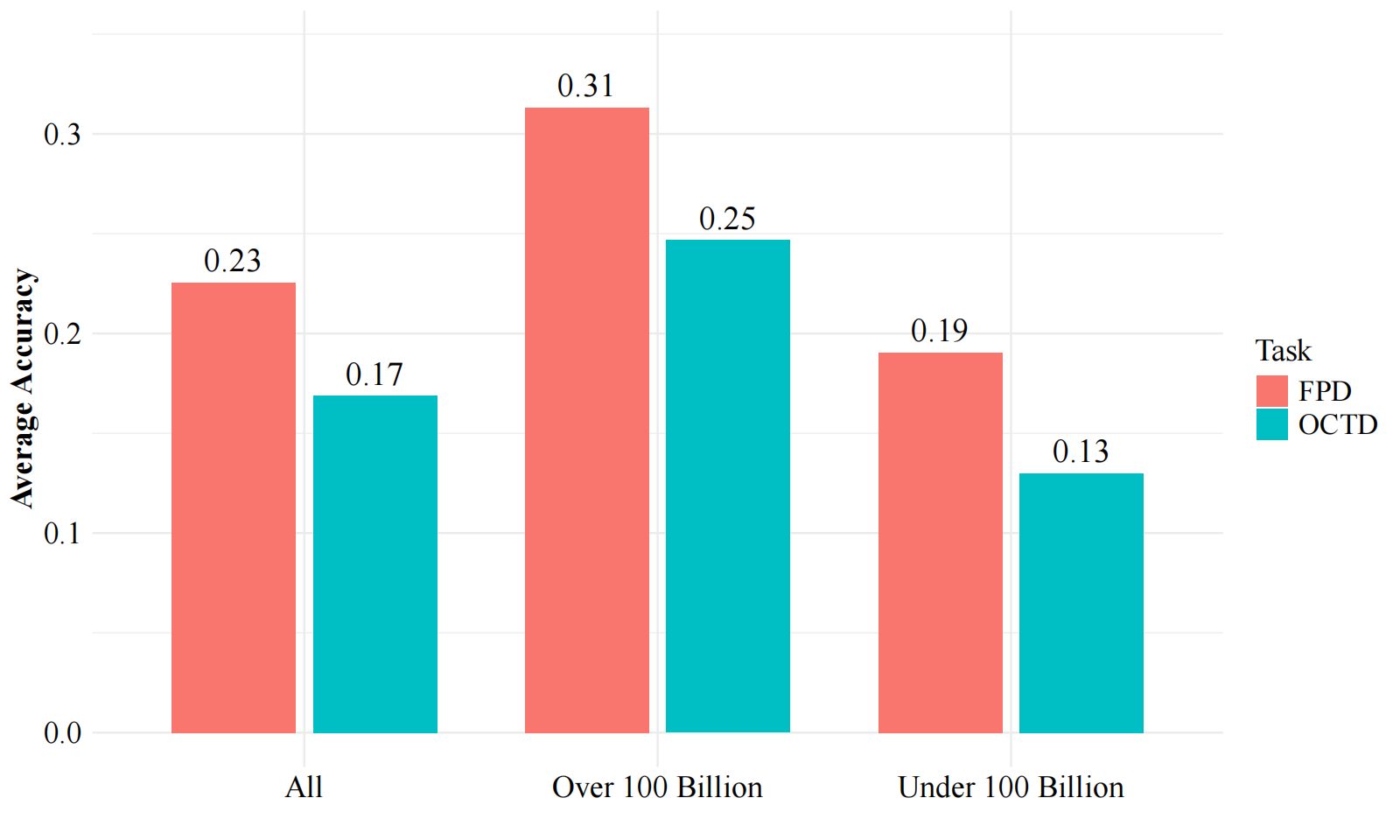}
    \caption{Average Accuracy of Different Parameters of MLLMs in FPD and OCTD}
    \raggedright
   Note: \textit{The bar chart illustrates the performance differences between models with over and less than 100 billion parameters on FPD and OCTD tasks. MLLMs, multimodal large language models. FPD, fundus photograph diagnosis. OCTD, optical coherence tomography diagnosis. }
\end{figure}
\subsection{Comparison of Cost and Benefit}
Due to the differences in model service providers, the prices for calling APIs of open-source large models also vary, so we do not make specific comparisons here. However, in the FPD task, it is evident that the accuracy of Qwen-2.5-VL-72B-Instruct is higher than that of GPT-4o (the most expensive). Among non-open-source large models, Gemini-2.0-Flash achieves the highest accuracy at the lowest price in both tasks.

\section{\textbf{\textbf{Discussion}}}

\subsection{Analysis for Results of MLLMs}
Our benchmark leverages multimodal data to evaluate the diagnostic accuracy of seven MLLMs for identifying seventeen different eye conditions using fundus photographs. Additionally, it assesses the performance of six MLLMs in detecting eight eye conditions based on OCT images. The results reveal significant accuracy differences among the models, indicating both the uneven capabilities of these models in handling multimodal data and the discriminative power of our benchmark in distinguishing their performance levels. Besides, the diagnostic accuracy of the same model varies significantly across different diseases, with some diseases even showing zero accuracy, such as hypertensive retinopathy (HTR), optic disc pallor (ODP), and choroidal neovascularization (CNV). These low-scoring diseases are commonly reflected across various models, indicating that current MLLMs struggle to accurately recognize fundus photographs or OCT images of certain diseases. This may be due to several factors. On one hand, the diagnosis of these diseases often involves overlapping or contained relationships. For example, CNV is a form of retinal neovascularization that can be caused by multiple fundus diseases\cite{sacconi2023towards}\cite{dansingani2015optical}\cite{velez2024long}. When presented with an image of CNV, the model may incorrectly identify other underlying diseases that could lead to CNV. On the other hand, the diagnosis of some diseases requires additional clinical context or examinations. For instance, high myopia-related macular degeneration necessitates a history of high myopia as diagnostic support, and retinal laser scars require a history of laser surgery. Particularly some diseases require a diagnosis of exclusion and are more dependent on complete clinical information and examinations\cite{li2022choriocapillaris}\cite{karst2019retinal}\cite{chen2022systematic}. In such cases, fundus photographs and OCT images alone can be ambiguous, making machine diagnosis particularly challenging. Moreover, the unequal distribution of data for each disease in the database also contributes to these challenges. Diseases with limited data may exhibit accidental errors in diagnosis, while those with larger datasets tend to have more stable and reliable accuracy.

In addition to the aforementioned results, the judgment of different models on normal fundus pictures or OCT images warrants special attention. The overall accuracy of the models in this regard is not high and varies, suggesting a bottleneck in the application of these models for disease screening\cite{jiang2017artificial}\cite{meng2024application}\cite{wu2024collaborative}. However, it is crucial to note that missing diseases during the screening process can lead to serious consequences and delay the optimal timing for treatment\cite{jeyaram2023unraveling}\cite{goh2024large}. Therefore, the repercussions of a machine misidentifying a pathological image as normal are potentially more severe than confusing pathological images. Conversely, mistaking a normal image for a pathological one can have even more serious consequences, as it not only reduces the accuracy of screening but also increases the cost of further examinations\cite{chustecki2024benefits}\cite{cabral2025future}. Hence, greater emphasis should be placed on the accuracy of machine recognition of normal images, and the large models discussed in this paper require further refinement.

Overall, our benchmark effectively evaluates the models' performance in processing multimodal information, largely due to our robust database construction methods. First, our data is sourced from open literature, where we identify and collect fundus or OCT images. These literature-derived images are more clinically relevant and carry a lower risk of data leakage compared to open medical exam question banks, making them both practical and challenging for large models. Second, we employ a rigorous multi-expert cross-verification process. Senior clinical ophthalmologists independently review and annotate the data, followed by a cross-check among experts to ensure diagnostic accuracy, eliminate duplicates, and remove ambiguous images. This collaborative and iterative approach enhances the reliability and robustness of our benchmark. Additionally, our benchmark utilizes restricted-domain answers, offering a broader range of options for models to reference. This approach reduces the randomness associated with single-choice questions and minimizes the likelihood of highly deviant answers, thereby better assessing the models' true capabilities. The significant variability in model scores demonstrates that our benchmark has strong differentiation power. It can identify current model weaknesses, pinpoint clinically challenging pain points based on low-scoring diseases, and suggest directions for further model optimization.
\subsection{Limitations}
However, we also identified several limitations of the current benchmark. First, the data is sourced from real clinical images, which, while beneficial for clinical relevance, inevitably introduces variability in image quality. Factors such as differences in equipment models, operator skill, environmental lighting, and patient cooperation can subtly affect image quality\cite{ott2022mapping}\cite{kaczmarczyk2024evaluating}. Increasing the volume of data can help enhance data stability, but these inherent quality differences remain a challenge. Second, relying solely on clinical images for diagnosis does not fully reflect real-world clinical practice. Many diseases require a comprehensive assessment that includes additional information such as patient complaints, medical history, surgical history, and family history. Unfortunately, publicly available resources with complete clinical data chains are scarce, likely due to patient privacy and ethical concerns\cite{shen2023chatgpt}\cite{li2023ethics}. This suggests the need to collect more comprehensive data in real clinical settings and establish a clinical multimodal information chain to support better diagnostic decisions by large models. Additionally, accurate disease staging is crucial for many conditions. For example, diabetic retinopathy presents differently at various stages, each requiring distinct management approaches\cite{cheung2010diabetic}\cite{kollias2010diabetic}. The current diagnostic accuracy of large models is insufficient for such nuanced tasks. By improving their accuracy, we can set higher and more detailed requirements for disease staging, ultimately enhancing the models' clinical utility.
\subsection{Future Work}
Our study offers valuable insights for the future development of multimodal benchmarks. First, combining multiple types of images may enhance model accuracy. In this study, we independently examined the models' judgments based on fundus photographs and OCT images. However, referencing both fundus photographs and OCT images simultaneously could potentially yield superior diagnostic accuracy compared to using a single type of image. This aligns with our earlier observation that integrating comprehensive clinical data can improve model performance. Future multimodal model development should emphasize the commonalities in the performance of different imaging methods for the same disease, making models more adaptable to real-world clinical scenarios. Furthermore, images are less influenced by social factors such as region and culture compared to language\cite{akinseloyin2024question}. This gives multimodal models a natural advantage in standardization and normalization over traditional large language models, thereby enhancing their global universality. However, greater attention should be paid to ensuring uniform image quality. Additionally, it has been noted in many previous studies that specialized vertical models have not outperformed generalized large language models in evaluations\cite{fast2024autonomous}. This suggests that current large model development still faces a bottleneck in enhancing in-depth clinical thinking. The challenges of real clinical scenarios may not be effectively addressed by specialized models alone. The efforts in ophthalmology highlighted in our study will contribute to the development of more specialized, refined, and in-depth models, potentially overcoming some of these limitations.
\bibliography{references.bib} 

\begin{thebibliography}{10}

\bibitem{yu2023leveraging}
P.~Yu, H.~Xu, X.~Hu, et~al.
\newblock Leveraging generative ai and large language models: A comprehensive roadmap for healthcare integration.
\newblock {\em Healthcare (Basel)}, 11(20), 2023.

\bibitem{zhao2021calibrate}
Z.~Zhao, E.~Wallace, and S.~Feng.
\newblock Calibrate before use: Improving few-shot performance of language models.
\newblock In {\em Proceedings of the 38th International Conference on Machine Learning}, Proceedings of Machine Learning Research, pages 12697--707. PMLR, 2021.

\bibitem{cascella2023evaluating}
M.~Cascella, J.~Montomoli, V.~Bellini, et~al.
\newblock Evaluating the feasibility of chatgpt in healthcare: An analysis of multiple clinical and research scenarios.
\newblock {\em J Med Syst}, 47(1):33, 2023.

\bibitem{li2024integrated}
J.~Li, Z.~Guan, J.~Wang, et~al.
\newblock Integrated image-based deep learning and language models for primary diabetes care.
\newblock {\em Nat Med}, 30(10):2886--2896, 2024.

\bibitem{jin2024matching}
Q.~Jin, Z.~Wang, C.~S. Floudas, et~al.
\newblock Matching patients to clinical trials with large language models.
\newblock {\em Nat Commun}, 15(1):9074, 2024.

\bibitem{shen2023chatgpt}
Y.~Shen, L.~Heacock, J.~Elias, et~al.
\newblock Chatgpt and other large language models are double-edged swords.
\newblock {\em Radiology}, 307(2):e230163, 2023.

\bibitem{zhou2024pre}
J.~Zhou, X.~He, L.~Sun, et~al.
\newblock Pre-trained multimodal large language model enhances dermatological diagnosis using {SkinGPT-4}.
\newblock {\em Nat Commun}, 15(1):5649, 2024.

\bibitem{bhattacharya2024large}
M.~Bhattacharya, S.~Pal, S.~Chatterjee, et~al.
\newblock Large language model to multimodal large language model: A journey to shape the biological macromolecules to biological sciences and medicine.
\newblock {\em Mol Ther Nucleic Acids}, 35(3):102255, 2024.

\bibitem{liu2023benchmarking}
J.~Liu, P.~Zhou, Y.~Hua, et~al.
\newblock Benchmarking large language models on {CMExam} - a comprehensive chinese medical exam dataset.
\newblock In {\em Proceedings of the 37th International Conference on Neural Information Processing Systems}, page Article 2283, New Orleans, LA, USA, 2023. Curran Associates Inc.

\bibitem{besler2024accuracy}
M.~S. Beşler.
\newblock The accuracy of the multimodal large language model {GPT-4} on sample questions from the interventional radiology board examination.
\newblock {\em Acad Radiol}, 31(8):3476, 2024.

\bibitem{alryalat2024evaluating}
S.~A. Alryalat, A.~M. Musleh, and M.~Y. Kahook.
\newblock Evaluating the strengths and limitations of multimodal {ChatGPT-4} in detecting glaucoma using fundus images.
\newblock {\em Front Ophthalmol (Lausanne)}, 4:1387190, 2024.

\bibitem{jin2023medcpt}
Q.~Jin, W.~Kim, Q.~Chen, et~al.
\newblock {MedCPT}: Contrastive pre-trained transformers with large-scale {PubMed} search logs for zero-shot biomedical information retrieval.
\newblock {\em Bioinformatics}, 39(11), 2023.

\bibitem{liu2024medbench}
M.~Liu, J.~Ding, J.~Xu, et~al.
\newblock {MedBench}: A comprehensive, standardized, and reliable benchmarking system for evaluating chinese medical large language models.
\newblock arXiv, 2024.

\bibitem{lee2006beyond}
M.~Lee, J.~Cimino, H.~R. Zhu, et~al.
\newblock Beyond information retrieval--medical question answering.
\newblock In {\em AMIA Annu Symp Proc}, pages 469--473, 2006.

\bibitem{liu2023medical}
F.~Liu, T.~Zhu, X.~Wu, et~al.
\newblock A medical multimodal large language model for future pandemics.
\newblock {\em NPJ Digit Med}, 6(1):226, 2023.

\bibitem{pal2022medmcqa}
A.~Pal, L.~K. Umapathi, and M.~Sankarasubbu.
\newblock {MedMCQA} : A large-scale multi-subject multi-choice dataset for medical domain question answering.
\newblock In {\em ACM Conference on Health, Inference, and Learning}, 2022.

\bibitem{lin2023comparison}
J.~C. Lin, D.~N. Younesi, S.~S. Kurapati, et~al.
\newblock Comparison of {GPT-3.5}, {GPT-4}, and human user performance on a practice ophthalmology written examination.
\newblock {\em Eye (Lond)}, 37(17):3694--3695, 2023.

\bibitem{healey2025llm}
E.~Healey and I.~Kohane.
\newblock {LLM-CGM}: A benchmark for large language model-enabled querying of continuous glucose monitoring data for conversational diabetes management.
\newblock {\em Pac Symp Biocomput}, 30:82--93, 2025.

\bibitem{chen2024evaluating}
X.~Chen, L.~Wang, M.~You, et~al.
\newblock Evaluating and enhancing large language models' performance in domain-specific medicine: Development and usability study with {DocOA}.
\newblock {\em J Med Internet Res}, 26:e58158, 2024.

\bibitem{ma2022benchmarking}
D.~Ma, M.~R.~H. Taher, J.~Pang, et~al.
\newblock Benchmarking and boosting transformers for medical image classification.
\newblock In {\em Domain Adaptation and Representation Transfer}, volume 13542, pages 12--22, 2022.

\bibitem{kus2024medsegbench}
Z.~Kuş and M.~Aydın.
\newblock {MedSegBench}: A comprehensive benchmark for medical image segmentation in diverse data modalities.
\newblock {\em Sci Data}, 11(1):1283, 2024.

\bibitem{gong2023pimedseg}
X.~Gong, L.~Wang, L.~Miao, et~al.
\newblock {PIMedSeg}: Progressive interactive medical image segmentation.
\newblock {\em Comput Methods Programs Biomed}, 241:107776, 2023.

\bibitem{kocak2025bias}
B.~Koçak, A.~Ponsiglione, A.~Stanzione, et~al.
\newblock Bias in artificial intelligence for medical imaging: fundamentals, detection, avoidance, mitigation, challenges, ethics, and prospects.
\newblock {\em Diagn Interv Radiol}, 31(2):75--88, 2025.

\bibitem{liu2020evolving}
B.~Liu, W.~Chi, X.~Li, et~al.
\newblock Evolving the pulmonary nodules diagnosis from classical approaches to deep learning-aided decision support: three decades' development course and future prospect.
\newblock {\em J Cancer Res Clin Oncol}, 146(1):153--185, 2020.

\bibitem{ting2019artificial}
D.~S.~W. Ting, L.~R. Pasquale, L.~Peng, et~al.
\newblock Artificial intelligence and deep learning in ophthalmology.
\newblock {\em Br J Ophthalmol}, 103(2):167--175, 2019.

\bibitem{costin2025artificial}
H.-N. Costin, M.~Fira, and L.~Goraș.
\newblock Artificial intelligence in ophthalmology: Advantages and limits.
\newblock {\em Applied Sciences}, 15(4):1913, 2025.

\bibitem{fujinami2019prediction}
Y.~Fujinami-Yokokawa, N.~Pontikos, L.~Yang, et~al.
\newblock Prediction of causative genes in inherited retinal disorders from spectral-domain optical coherence tomography utilizing deep learning techniques.
\newblock {\em J Ophthalmol}, 2019:1691064, 2019.

\bibitem{li2023generic}
H.~Li, H.~Liu, H.~Fu, et~al.
\newblock A generic fundus image enhancement network boosted by frequency self-supervised representation learning.
\newblock {\em Med Image Anal}, 90:102945, 2023.

\bibitem{baxter2022data}
S.~L. Baxter, K.~Nwanyanwu, G.~Legault, et~al.
\newblock Data sources for evaluating health disparities in ophthalmology: Where we are and where we need to go.
\newblock {\em Ophthalmology}, 129(10):e146--e149, 2022.

\bibitem{lee2016evaluating}
C.~S. Lee, A.~Morris, R.~N. Van~Gelder, et~al.
\newblock Evaluating access to eye care in the contiguous united states by calculated driving time in the united states medicare population.
\newblock {\em Ophthalmology}, 123(12):2456--2461, 2016.

\bibitem{sacconi2023towards}
R.~Sacconi, S.~Fragiotta, D.~Sarraf, et~al.
\newblock Towards a better understanding of non-exudative choroidal and macular neovascularization.
\newblock {\em Prog Retin Eye Res}, 92:101113, 2023.

\bibitem{dansingani2015optical}
K.~K. Dansingani and K.~B. Freund.
\newblock Optical coherence tomography angiography reveals mature, tangled vascular networks in eyes with neovascular age-related macular degeneration showing resistance to geographic atrophy.
\newblock {\em Ophthalmic Surg Lasers Imaging Retina}, 46(9):907--912, 2015.

\bibitem{velez2024long}
R.~Velez-Montoya, H.~K. Osorio-Landa, K.~C. Franco-Ramirez, et~al.
\newblock Long-term functional, anatomical outcome, and qualitative analysis by {OCTA}, as a predictor of disease recurrences in patients with choroidal neovascularization secondary to angioid streaks.
\newblock {\em Int J Retina Vitreous}, 10(1):53, 2024.

\bibitem{li2022choriocapillaris}
J.~Li, H.~Zhou, M.~Feinstein, et~al.
\newblock Choriocapillaris changes in myopic macular degeneration.
\newblock {\em Transl Vis Sci Technol}, 11(2):37, 2022.

\bibitem{karst2019retinal}
S.~G. Karst, H.~Beiglboeck, R.~Scharinger, et~al.
\newblock Retinal and choroidal perfusion status in the area of laser scars assessed with swept-source optical coherence tomography angiography.
\newblock {\em Invest Ophthalmol Vis Sci}, 60(14):4865--4871, 2019.

\bibitem{chen2022systematic}
Y.~Chen, X.~Han, I.~Gordon, et~al.
\newblock A systematic review of clinical practice guidelines for myopic macular degeneration.
\newblock {\em J Glob Health}, 12:04026, 2022.

\bibitem{jiang2017artificial}
F.~Jiang, Y.~Jiang, H.~Zhi, et~al.
\newblock Artificial intelligence in healthcare: past, present and future.
\newblock {\em Stroke Vasc Neurol}, 2(4):230--243, 2017.

\bibitem{meng2024application}
X.~Meng, X.~Yan, K.~Zhang, et~al.
\newblock The application of large language models in medicine: A scoping review.
\newblock {\em iScience}, 27(5):109713, 2024.

\bibitem{wu2024collaborative}
S.~H. Wu, W.~J. Tong, M.~D. Li, et~al.
\newblock Collaborative enhancement of consistency and accuracy in us diagnosis of thyroid nodules using large language models.
\newblock {\em Radiology}, 310(3):e232255, 2024.

\bibitem{jeyaram2023unraveling}
M.~Jeyaraman, S.~Balaji, N.~Jeyaraman, et~al.
\newblock Unraveling the ethical enigma: Artificial intelligence in healthcare.
\newblock {\em Cureus}, 15(8):e43262, 2023.

\bibitem{goh2024large}
E.~Goh, R.~Gallo, J.~Hom, et~al.
\newblock Large language model influence on diagnostic reasoning: A randomized clinical trial.
\newblock {\em JAMA Netw Open}, 7(10):e2440969, 2024.

\bibitem{chustecki2024benefits}
M.~Chustecki.
\newblock Benefits and risks of ai in health care: Narrative review.
\newblock {\em Interact J Med Res}, 13:e53616, 2024.

\bibitem{cabral2025future}
B.~P. Cabral, L.~A.~M. Braga, C.~G. Conte~Filho, et~al.
\newblock Future use of ai in diagnostic medicine: 2-wave cross-sectional survey study.
\newblock {\em J Med Internet Res}, 27:e53892, 2025.

\bibitem{ott2022mapping}
S.~Ott, A.~Barbosa-Silva, K.~Blagec, et~al.
\newblock Mapping global dynamics of benchmark creation and saturation in artificial intelligence.
\newblock {\em Nat Commun}, 13(1):6793, 2022.

\bibitem{kaczmarczyk2024evaluating}
R.~Kaczmarczyk, T.~I. Wilhelm, R.~Martin, et~al.
\newblock Evaluating multimodal ai in medical diagnostics.
\newblock {\em NPJ Digit Med}, 7(1):205, 2024.

\bibitem{li2023ethics}
H.~Li, J.~T. Moon, S.~Purkayastha, et~al.
\newblock Ethics of large language models in medicine and medical research.
\newblock {\em Lancet Digit Health}, 5(6):e333--e335, 2023.

\bibitem{cheung2010diabetic}
N.~Cheung, P.~Mitchell, and T.~Y. Wong.
\newblock Diabetic retinopathy.
\newblock {\em Lancet}, 376(9735):124--136, 2010.

\bibitem{kollias2010diabetic}
A.~N. Kollias and M.~W. Ulbig.
\newblock Diabetic retinopathy: Early diagnosis and effective treatment.
\newblock {\em Dtsch Arztebl Int}, 107(5):75--83, 2010.

\bibitem{akinseloyin2024question}
O.~Akinseloyin, X.~Jiang, and V.~Palade.
\newblock A question-answering framework for automated abstract screening using large language models.
\newblock {\em J Am Med Inform Assoc}, 31(9):1939--1952, 2024.

\bibitem{fast2024autonomous}
D.~Fast, L.~C. Adams, F.~Busch, et~al.
\newblock Autonomous medical evaluation for guideline adherence of large language models.
\newblock {\em NPJ Digit Med}, 7(1):358, 2024.

\end{thebibliography}
\bibliographystyle{IEEEtran} 

\end{document}